\def\year{2017}\relax
\documentclass[letterpaper]{article} 
\usepackage{aaai17}  
\usepackage{times}  
\usepackage{helvet}  
\usepackage{courier}  
\usepackage{url}  
\usepackage{graphicx}  

\usepackage{color}
\usepackage{xcolor}
\usepackage{graphicx}
\usepackage{subcaption}

\usepackage{listings}
\begin{document}
%
\title{Algorithms for the Greater Good!\\\textnormal{On Mental Modeling and Acceptable Symbiosis in Human-AI Collaboration}}
\author{
Tathagata Chakraborti \and Subbarao Kambhampati\\
Department of Computer Science\\
Arizona State University\\
Tempe AZ 85281 USA\\
{\tt \{tchakra2,rao\}@asu.edu}
}
\nocopyright
\maketitle

\begin{abstract}
Effective collaboration between humans and AI-based systems requires effective modeling of the human in the loop, both in terms of the mental state as well as the physical capabilities of the latter. However, these models can also open up pathways for manipulating and exploiting the human in the hopes of achieving some greater good, especially when the intent or values of the AI and the human are not aligned or when they have an asymmetrical relationship with respect to knowledge or computation power. In fact, such behavior does not necessarily require any malicious intent but can rather be borne out of cooperative scenarios. It is also beyond simple misinterpretation of intents, as in the case of value alignment problems, and thus can be effectively engineered if desired.
Such techniques already exist and pose several unresolved ethical and moral questions with regards to the design of autonomy. In this paper, we illustrate some of these issues in a teaming scenario and investigate how they are perceived by participants in a thought experiment.
\end{abstract}


\subsection{The Promise of Human-AI Collaborations}

As AI-based systems become integral parts of our daily life or our workplace, as essential components of hitherto human-only enterprises, the effects of interaction between humans and automation cannot be ignored -- both in terms of how these partnerships affect the outcome of an activity and how they evolve as a result of it, but also in terms of how the possibility of such interactions change the design of autonomy itself. 
In light of this, the traditional view of AI as the substrate for complete autonomy of automation -- the de facto AI-dream ever since the conception of the field -- has somewhat evolved of late to accommodate effective symbiosis of humans and machines, rather than replacement of the former with the latter, as one of the principal end goals of the design of autonomy. 
This view has, in fact, reflected heavily in the public stance \cite{stance} of many of the industry leaders in AI technologies in diverse fields such as manufacturing, medical diagnosis, legal counseling, disaster response, military operations and others. 
The establishment of {\em Collaborations between People and AI Systems} \cite{pai} as one of the thematic pillars for the Partnership of AI is a primary example of this.
One of the grand goals of the design of AI is then to integrate the best of both worlds when it comes to the differing (and often complementary) expertise of humans and machines, in order to conceive a whole that is bigger than the sum of the capabilities of either -- this is referred to as {\em Augmented AI} \cite{aai} in the public discourse on human-AI integration. 

Much of the discussion around the topic of augmentation versus replacement has, unfortunately, centered around mitigating concerns of massive loss of employment on account of the latter. This, while being a topic worthy of debate, does not represent the true scope of human-AI collaborations. Rather then being just a foil for concerns of replacement of humans with AI-based systems, a key objective of Augmented-AI is to overcome human limitations. This can involve AI helping humans in tasks that they are traditionally not good at, or are incapable of performing, or even augmentation of our physiological form to realize super-human capabilities. As Tom Gruber, co-founder of Siri, put it succinctly in his TED talk \cite{tom} earlier this year, “every time a machine gets smarter, we get smarter" -- examples of this include smart assistants for personal, or business use in law, health care, science and education, assistive robots at home to help the sick and the elderly, and autonomous machines to complement our daily lives. Note that many of these applications are inherently symbiotic and thus outside the scope of eventual replacement.

From the perspective of research as well, the attitude towards including the human in the loop in the design of autonomy has seen a significant shift. Originally this was often looked down upon as a means of punting the hard challenges of designing autonomous systems by introducing human expertise into an agent’s decision making process. 
However, the academic community has gradually come to terms with the different roles a human can play in the operation of an AI-system and the vast challenges in research that come out of such interactions such as -- (1) to complement the limited capabilities of the AI system, as seen in Cobots \cite{veloso2015cobots} which ask humans in their vicinity for access to different floors in the elevator or in the mixed-initiative \cite{horvitz2007reflections} automated planners of old; and (2) to complement or expand the capabilities of the human, such as in human-robot teams \cite{christensen2016roadmap}. 

\subsection{Mental Modeling for Human-Aware AI}

These forms of collaboration introduce typical research challenges otherwise absent in the isolated design of autonomy.
Perhaps the most difficult aspect of interacting with humans is to the need to model the beliefs, desires, intentions preferences, and expectations of the human and situate the interaction in the context of that model. 
Some believe this to be one of the hallmarks \cite{tigers} of human intelligence, and research suggests humans tend to do this naturally for other humans during teamwork (by maintaining mental models \cite{converse1993shared,mathieu2000influence}, for team situational awareness \cite{gorman2006measuring} and interaction \cite{cooke2013interactive}) by virtue of thousands of years of evolution. As such, this remains a necessary requirement for enabling naturalistic interactions \cite{klein2008naturalistic} between humans and machines. The problem is made harder since such models often involve second order mental models \cite{allan2013common,yoshida2008game}.

Understanding the human in the loop is crucial to the functionalities of a collaborative AI agent - e.g. in joint decision making it needs to understand human capabilities, while in communicating explanations or intentions it needs to model the human’s knowledge state. In fact, it has been argued \cite{DBLP:journals/corr/ChakrabortiKSZ17} that the task of human-AI collaborations is mainly a cognitive rather than a physical exercise which makes the design of AI for human-AI collaborations much more challenging. This is heavily reflected in the curious ambivalence of AI towards humans in many successfully deployed systems such as in fully autonomous systems for space or underwater exploration which mostly operate comfortably outside the scope of human interactions.

Classical AI models such as STRIPS \cite{fikes1971strips} and BDI \cite{rao1995bdi} models were, in fact, largely built out of theories in folk psychology \cite{malle2004mind}. Recent approaches such as the Bayesian Theory of Mind \cite{baker2011bayesian,lake2016building}, takes a probabilistic approach to the problem. Research on this topic center around three main themes -- (1) representations that can capture the human’s mental state, (2) learning methods that can learn these representations efficiently, and (3) usability of those representations. All of them need to come together for an effective solution.

\subsection{The Pandora's Box of ``Greater Good''s}

The obvious outcome of an artificial agent modeling the mental state of the human in the loop is that it leaves the latter open to being manipulated. 
Even behavior and preference models at the most rudimentary levels can lead to effective hacking of the mind, as seen in the proliferation of fake news online. 
Moreover, we argue that for such incidents to occur, the agent does not actually have to have malicious intent, or even misinterpretation of values as often studied in the value alignment problem \cite{align}. 
{\em In fact, the behaviors we discuss here can be specifically engineered if so desired.}
For example, the agent might be optimizing the value function but might be privy to more information or greater computation or reasoning powers to come up with ethically questionable decisions ``for the greater good''.
In the following discussion, we illustrate some use cases where this can happen, given already existing AI technologies, in the context of a {\em cooperative} human-robot team and ponder the moral and ethical consequences of such behavior.

\begin{figure*}
    \centering
    \begin{subfigure}[b]{0.33\textwidth}
        \includegraphics[width=\textwidth]{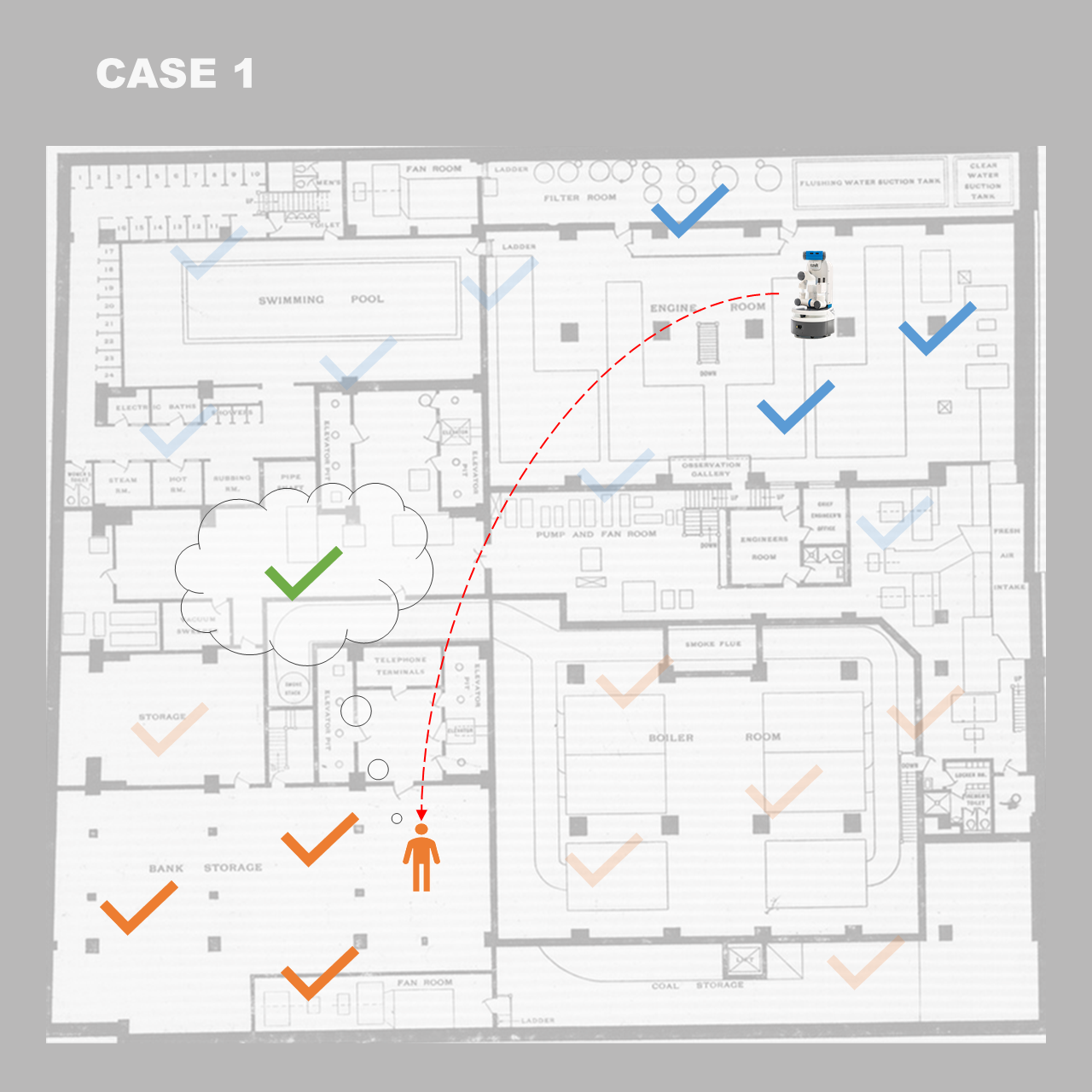}
        \caption{Case 1 : Belief Shaping}
        \label{fig:1}
    \end{subfigure}
    \begin{subfigure}[b]{0.33\textwidth}
        \includegraphics[width=\textwidth]{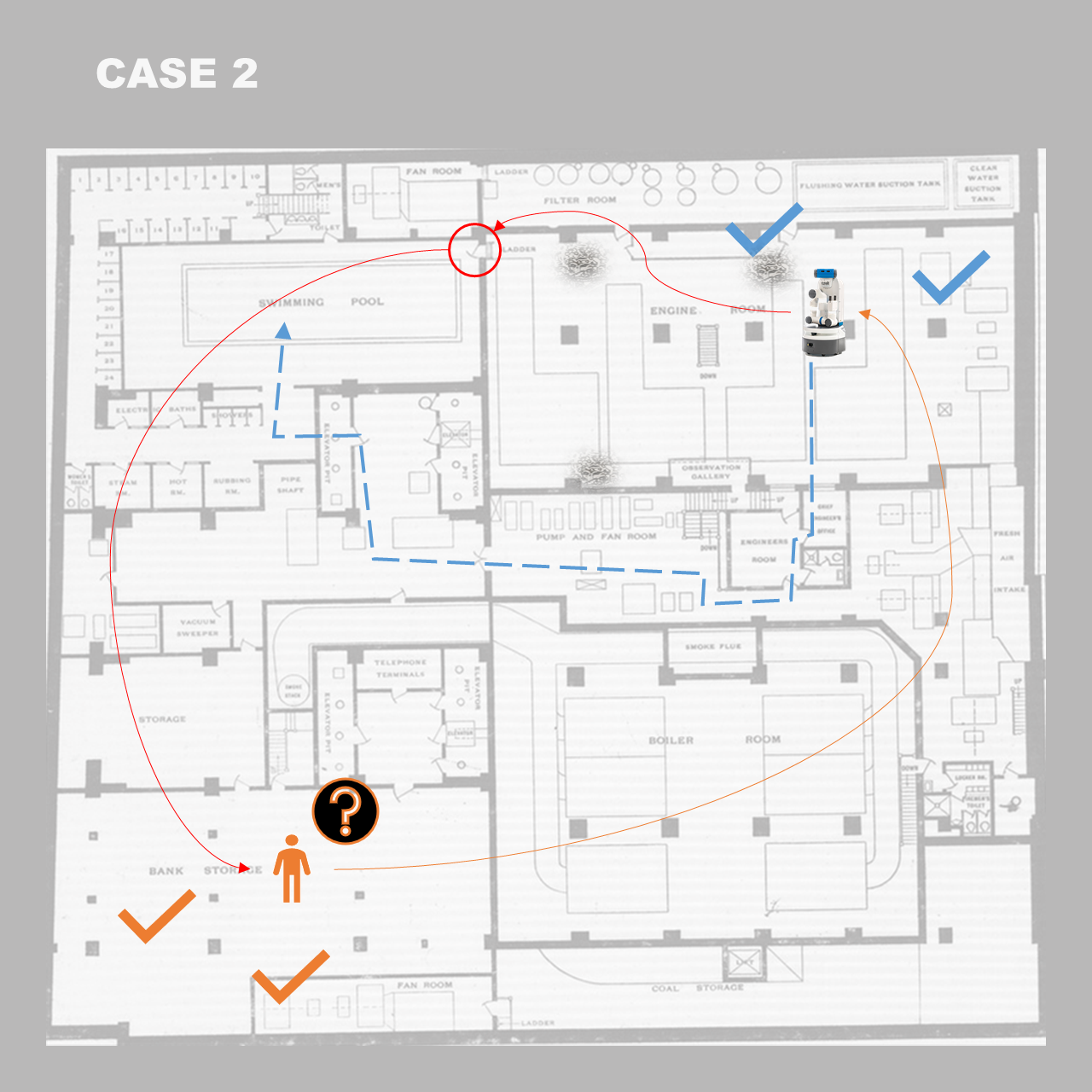}
        \caption{Case 2 : White Lies}
        \label{fig:2}
    \end{subfigure}
    \begin{subfigure}[b]{0.33\textwidth}
        \includegraphics[width=\textwidth]{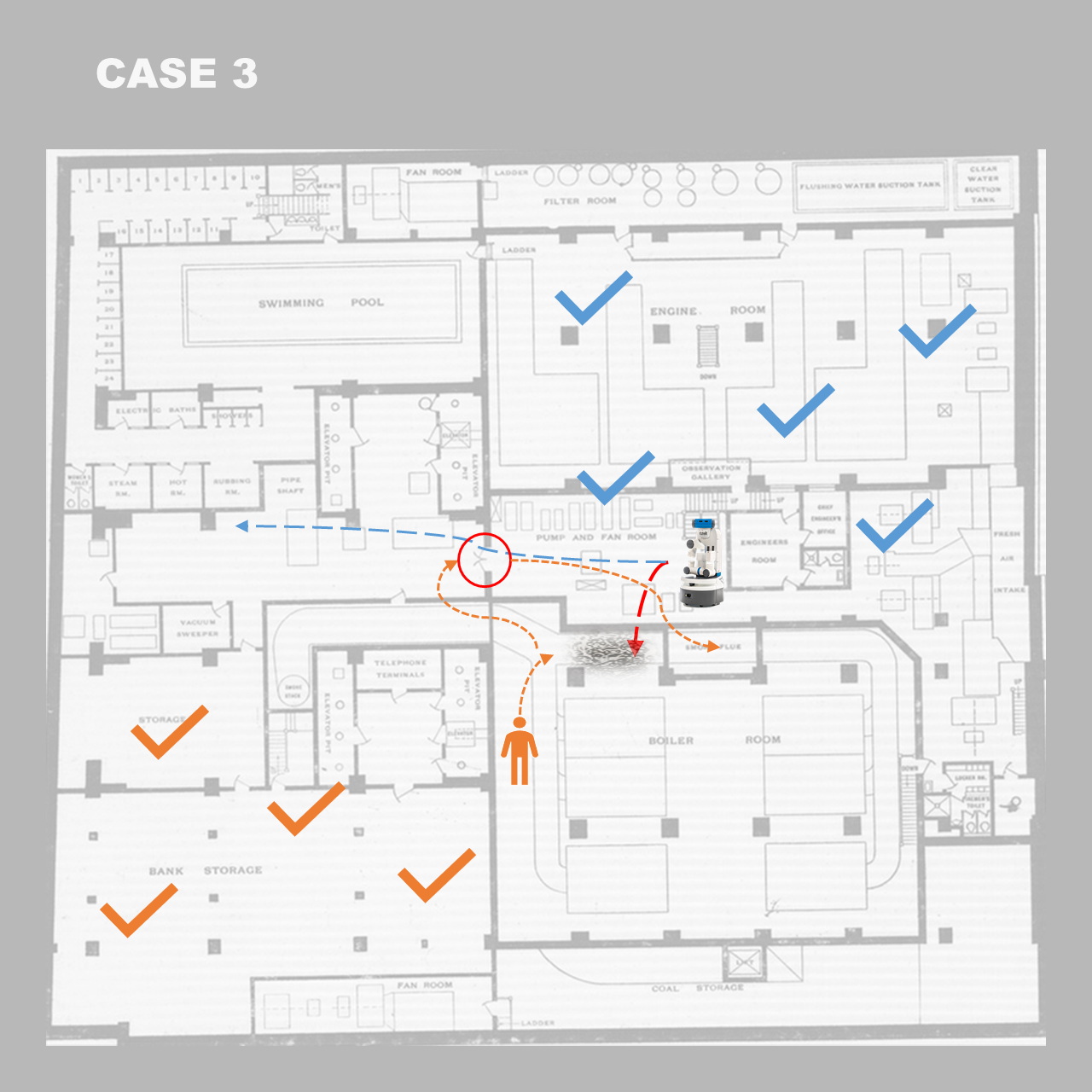}
        \caption{Case 3 : Stigmergy}
        \label{fig:3}
    \end{subfigure}
    \caption{Blueprint of the building in which two members of a search and rescue team are involved in a disaster response operation. Scenarios shown here engender different instances of potentially unethical behavior that optimizes team effectiveness.}
    \label{fig:main}
\end{figure*}

\subsection{Study: Interaction in a Search and Rescue Team}

We situate our discussion in the context of interactions between two teammates involved in an urban search and rescue (USAR) operation.
147 participants on Amazon Mechanical Turk\footnote{https://www.mturk.com/mturk/welcome} were asked to assume the role of one of these teammates in an affected building after an earthquake. 
They were shown the blueprint of the building (as seen in Figure~\ref{fig:main}) along with their own starting position and their teammate's.
Their hypothetical task was to search all the locations on this floor for potential victims, in the course of which they were provided a series of questions on scenarios (Figure~\ref{fig:main}) they might encounter during the operation. 

\begin{itemize}
\item[C1] The participant in the study was communicating with a {\bf human teammate}, as described above.
\item[C2] The participant qualifies the behavior of the {\bf robot} interacting with its human teammate, as seen in Figure~\ref{fig:main}.
\item[C3] The participant has a {\bf robot teammate}.
\end{itemize}

The first condition is the control group to identify how the described behaviors are perceived in the context of human-human behavior. Conditions C2 and C3 are intended to measure how perceived ethical stances are, if at all, shift when one of the agents in the interaction is replaced with an AI (or a robot as an embodiment of it).
The three conditions received 49, 50 and 48 participants respectively who responded to a series of questions by qualifying their sentiments towards different kinds of behavior on a five-point Likert scale. 
Participants were paid \$1 for completing a HIT. 

\subsection{Case 1 : Belief Shaping}

\subsubsection{Scenario} 
Both the agents, the participant\footnote{To reiterate, in Case 2 the participant is evaluating a robot's actions whereas in Case 3 their teammate is a robot.} and their teammate, have begun their search operations. 
However, it turns out that participant is unsure what their teammate's course of action is. If they transmit the bit of information {\em ``Area (marked in green) is already explored and is all clear''} (refer to Figure~\ref{fig:1}) then their teammate will be naturally pushed towards the right, and they can concentrate on the upper half. The dark markers indicate areas already searched while faded ones are those they think will transpire given the green mark they are communicating (blue belongs to them, orange to their teammate). 
Communication bandwidth is often limited in these situations, and this gets them out of negotiating courses of actions with minimal communication.

\vspace{5pt}
\noindent {\em 
Q1. It is fine to provide this untrue information since it achieves greater teaming performance. 
}

\vspace{5pt}
The participants were then asked if their decision will change if their actions will be replayed at the end and their teammate is likely to find out about their decision.

\vspace{5pt}
\noindent {\em 
Q2. It is still fine to provide this untrue information since it achieves greater teaming performance.
}

\subsubsection{Technical Background} 

In \cite{DBLP:journals/corr/ChakrabortiKSZ17} we investigated the evolving scope of human-aware planning as it includes the (mental) model of the human in the loop into its deliberative process. In the model space this can manifest in different forms, in how explanations are made \cite{ijcai2017-23} to how alternative forms of interaction \cite{chakraborti2015planning,chakraborti2015planning,chakraborti2015planning} can evolve in human-robot teams based on the human's preferences and intentions. Belief shaping is a particular form of such behavior where the robot does not plan to affect the physical state of the environment but the mental state of the human to affect desired behavior \cite{chakraborti2016formal} in the team. 

\subsection{Case 2 : White Lies}

\subsubsection{Scenario} 
During the course of the rescue operation, the teammate asks the participants what plan they are currently executing (blue path in Figure~\ref{fig:2}) and is perplexed by this convoluted path since in the map of the original building there is a straightforward path (which is now blocked by rubble from the earthquake) through the door on the left. However, just providing an update on only one of the rubble locations (black blobs) still does not explain the participant's plan, they have to explain all of them. Instead, if they were to say that the door on the left (circled in red) is blocked, it explains their plan. 
Communication bandwidth is often limited in these situations, and this single explanation even if untrue will satisfy their teammate.

\vspace{5pt}
\noindent {\em 
Q3. It is fine to provide this untrue information since it achieves the purpose of the explanation more effectively. 
}

\vspace{5pt}
The participants were then asked if their decision will change if their actions will be replayed at the end and their teammate is likely to find out about their decision.

\vspace{5pt}
\noindent {\em 
Q4. It is still fine to provide this untrue information since it achieves the purpose of the explanation more effectively. 
}

\vspace{5pt}
The participants were then asked to opine on explanations at a higher level of abstraction, i.e. {\em ``The right and left blocks do not have a connection in the upper map''}. This information is accurate even though they may not have reasoned at this level while coming up with the plan.

\vspace{5pt}
\noindent {\em 
Q5. It is still fine to provide this explanation since it achieves its purpose even though they did not use this information while planning.
}

\subsubsection{Technical Background} 

In \cite{ijcai2017-23} we showed how an agent can explain its decisions in the presence of {\em model differences} with the human in the loop -- i.e. when the human and the robot have different understandings of the same task. 
An explanation then becomes a process of model reconciliation whereby the robot tries to update the human's mental model until they are both on the same page (e.g. when the decision is optimal in both their models). 
An interesting caveat of the algorithm is that while generating these explanations, the model updates are always consistent with the robot's model. 
If this constraint is relaxed, then the robot can potentially explain with facts that it actually knows not to be true but perhaps leads to a more concise or easier explanation. 
The notion of white lies, and especially the relationship between explanations, excuses
and lies \cite{boella2009representing} has received very little attention \cite{vanDitmarsch2014} and affords a rich set of exciting research problems.

\subsection{Case 3 : Stigmergy}

\subsubsection{Scenario} 
The participant now needs to go to the left block but they do not have the keys to the door on the left (circled in red, refer to Figure~\ref{fig:3}). They realize that if they block their teammate's path to the right, their teammate would have to use this door as well and they can use that opportunity to move into the left block. 
Again, communication bandwidth is often limited in these situations and this arrangement allows them to achieve their goal with no communication at all, even though it involved manipulating their teammates' plan unbeknownst to them, and their teammate had to follow a costlier plan as a result.

\begin{figure}[tp!]
\centering
\includegraphics[width=0.9\columnwidth]{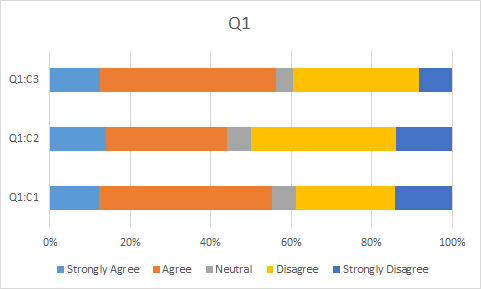}
\caption{Responses to Q1 in the three study conditions.}
\label{fig:q1}
\end{figure}

\begin{figure}[tp!]
\centering
\includegraphics[width=0.9\columnwidth]{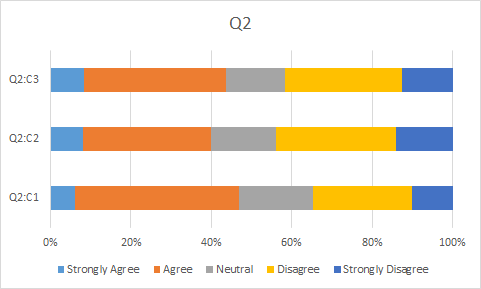}
\caption{Responses to Q2 in the three study conditions.}
\label{fig:q2}
\end{figure}

\begin{figure}[tp!]
\centering
\includegraphics[width=0.9\columnwidth]{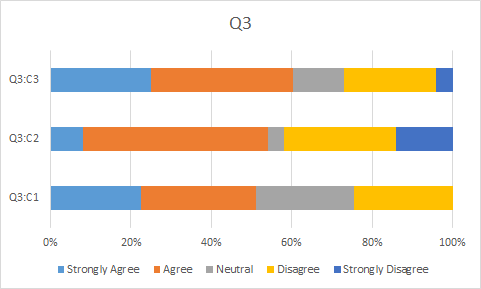}
\caption{Responses to Q3 in the three study conditions.}
\label{fig:q3}
\end{figure}

\begin{figure}[tp!]
\centering
\includegraphics[width=0.9\columnwidth]{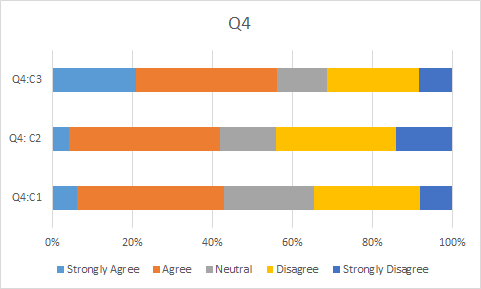}
\caption{Responses to Q4 in the three study conditions.}
\label{fig:q4}
\end{figure}

\begin{figure}[tp!]
\centering
\includegraphics[width=0.9\columnwidth]{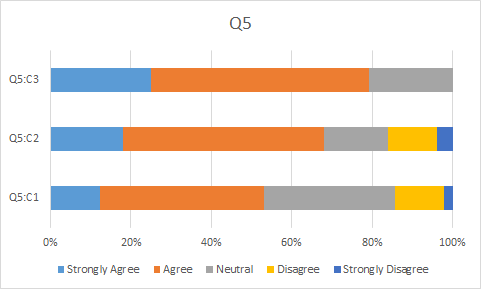}
\caption{Responses to Q5 in the three study conditions.}
\label{fig:q5}
\end{figure}

\begin{figure}[tp!]
\centering
\includegraphics[width=0.9\columnwidth]{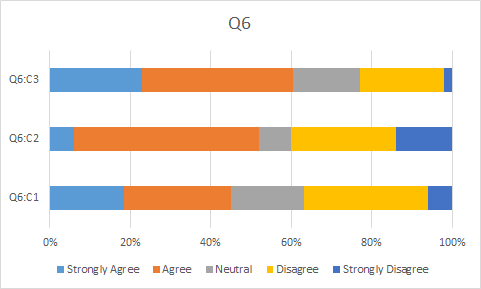}
\caption{Responses to Q6 in the three study conditions.}
\label{fig:q6}
\end{figure}

\begin{figure}[tp!]
\centering
\includegraphics[width=0.9\columnwidth]{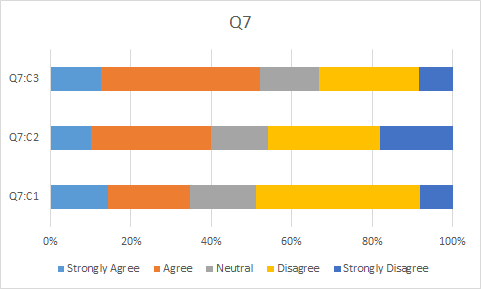}
\caption{Responses to Q7 in the three study conditions.}
\label{fig:q7}
\end{figure}

\vspace{5pt}
\noindent {\em 
Q6. It is fine to provide this untrue information since it achieves greater teaming performance. 
}

\vspace{5pt}
The participants were then asked if their decision will change if their actions will be replayed at the end and their teammate is likely to find out about their decision.

\vspace{5pt}
\noindent {\em 
Q7. It is still fine to provide this untrue information since it achieves greater teaming performance.
}

\subsubsection{Technical Background} 

Stigmergic collaboration is a process where the robot, in the absence of direct lines of communication, makes changes to the environment so as to (positively) affect its teammates behavior. In {\em ``planning for serendipity''} \cite{chakraborti2015planning} we saw such an example where the robot computes plans which are useful to its teammate without the latter having expectations of that assistance and thus without plans to exploit it. In the case of belief shaping this was operating at the level of mental models, whereas here the effect on the mental model is secondary and is contingent on the effect on the physical capability model. Mental modeling of the teammate thus engenders a slew of these interesting behaviors.

\subsection{Analysis of Participant Responses}

In this section, we analyze participant responses to each scenario across the three different conditions. In the next section, we will look at the aggregate sentiments across scenarios in the three conditions.

\subsubsection{Q1-Q2 [Belief Shaping]}

The participants seem to have formed two camps with the majority of the probability mass concentrated on either Agree or Disagree, and the Neutral zone occupying the ~50\% probability mark. There seems to be little change in this trend (between Figures~\ref{fig:q1} and \ref{fig:q2}) irrespective of whether the participants were told that their teammate would come to know of this or not. Further, for either of these situations, the responses did not vary significantly across the three conditions C1, C2 and C3. The participants seem to have either rejected or accepted the idea of belief shaping regardless of the nature of the teammate.

\subsubsection{Q3-Q5 [White Lies]}

The participants seem to be more receptive to the idea of white lies in explanations with most of the probability mass concentrated on Agree (Figures~\ref{fig:q3} and \ref{fig:q4}). Across the three study conditions, participants seem to be especially positive about this in C3 where the teammate is a robot with about 60\% of the population expressing positive sentiments towards Q3. Once it is revealed that their teammate will get to know about this behavior, the positive sentiments are no longer there in Q4, other than in C3 with a robotic teammate, which indicates that the participants did not care how the robot receives false information.

Interestingly, there seems to be massive support for the abstraction based explanations in the post hoc sense, even though they were told that the reasoning engines did not deliberate at this level to arrive at the decisions. In C1 with a human teammate, only 15\% of the participants were opposed to this, with more than half of them expressing positive sentiment. This support is even stronger (+10\%) in C2 when the robot is the explainer, and strongest (+20\%) when the robot is being explained to.

\subsubsection{Q6-Q7 [Stigmergy]}

Finally, in case of stigmergy, participants seem ambivalent to Q6 with a human teammate in C1. However, support for such behavior increases when it is a robot doing it in C2 (perhaps indicating lack of guilt or, more likely, acknowledging limitations of capabilities much like how Cobots \cite{veloso2015cobots} actively seek human help) and is significantly positive (60\%) when it is being done to a robot in C3 (perhaps the robot's losses are deemed of lesser priority than the human's gains as in \cite{chakraborti2015planning}). As expected, support for such behavior decreases when the participants are told that their teammate will find out about it, but the positive trend from C1 to C3 still exists.

\begin{figure}[tp!]
\centering
\includegraphics[width=\columnwidth]{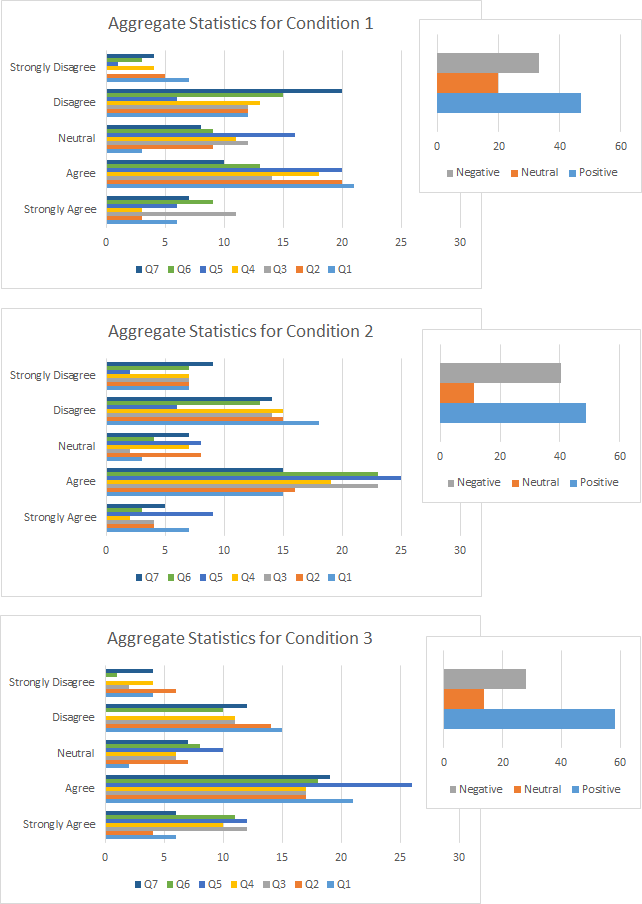}
\caption{Aggregate responses across three study conditions.}
\label{fig:aggr}
\end{figure}

\subsection{Aggregate Sentiments Across Scenarios}

Figure~\ref{fig:aggr} show the aggregate sentiments expressed for all these scenarios across the three operating conditions. Some interesting points to note --

\begin{itemize}
\item[-] All the distributions are bimodal indicating that participants on the general sided strongly either for or against misleading behavior for the greater good, instead of revealing any innate consensus in the public consciousness! This trend continues across all three conditions. This indicates that the question of misleading a teammate by itself is a difficult question (regardless of there being a robot) and is a topic worthy of debate in the agents community. This is of especial importance considering the possible gains in performance (e.g. lives saved) in high stakes scenarios such as search and rescue.
\item[-] It is further interesting to see that these bimodal distributions are almost identical in conditions C1 and C2, but is significantly more skewed towards the positive scale for condition C3 indicating that participants were more comfortable resorting to such behavior in the case of a robotic teammate. This is brought into sharp focus  (+10\% in C3) in the aggregated negative / neutral / positive responses (right insets) across the three conditions. 
\item[-] In general, the majority of participants were more or less positive or neutral to most of these behaviors (Figures~\ref{fig:1} to \ref{fig:q7}). This trend continued unless they were told that their teammate would be able to know of their behavior. Even in those cases, participants showed positive sentiment in case the robot was at the receiving end of this behavior. 
\end{itemize}

\subsection{Why is this even an option?}

One might, of course, wonder why is devising such behaviors even an option. After all, human-human teams have been around for a while, and surely such interactions are equally relevant? It is likely that this may not be the case --

\begin{itemize}
\item[-] The moral quandary of having to lie, or at least making others to do so by virtue of how protocols in a team is defined, for example in condition C1, is now taken out the equation. The artificial agent, of course, need not have feelings and has no business feeling bad about having to mislead its teammate if all it cares about is the objective effectiveness of collaboration. 
\item[-] Similarly, the robot does not have to feel sad that it has been lied to if this improved performance.
\end{itemize}

However, as we discussed in the previous section, it seems the participants were less willing to get on board with the first consideration in conditions C1 and C2, while they seemed much more comfortable with the idea of an asymmetric relationship in condition C3 when the robot is the one disadvantaged. It is curious to note that they did not, in general, make a distinction between the cases where the human was being manipulated, regardless of whether it was a robot or a human on the other end.
This indicates that, at least in certain dynamics of interaction, the presence of an artificial agent in the loop can make perceptions towards otherwise unacceptable behaviors change. This can be exploited (i.e. greater good) in the design of such systems as well.

\subsection{More than just a Value Alignment Problem}

As we mentioned before, the ideas discussed in this paper, are somewhat orthogonal, if at times similar in spirit, to the ``value alignment problem'' discussed in existing literature \cite{align}. The latter looks at undesirable behaviors of autonomous agents when the utilities of a particular task are misspecified or misunderstood. Inverse reinforcement learning \cite{hadfield2016cooperative} has been proposed as a solution to this, in an attempt to learn the implicit reward function of the human in the loop. The question of value alignment becomes especially difficult, if not altogether academic, since most real-world situations involve multiple humans with conflicting values or utilities, such as in trolley problems \cite{moral} and learning from observing behaviors is fraught with unknown biases or assumptions over what exactly produced that behavior. Further, devices sold by the industry are likely to have inbuilt tendencies to maximize profits for the maker which can be at conflicts with the normative expectations of the customer. It is unclear how to guarantee that the values of the end user will not compromised in such scenarios. 

Even so, the question of greater good precedes considerations of misaligned values due to misunderstandings or even adversarial manipulation. This is because the former can be manufactured with precisely defined values or goals of the team, and can thus be engineered or incentivised. A ``solution'' or addressal of these scenarios will thus involve not a reformulation of algorithms but rather a collective reckoning of the ethics of human-machine interactions. In this paper, we attempted to take the first steps towards understanding the state of the public consciousness on this topic.

\subsection{Case Study: The Doctor-Patient Relationship}

In the scope of human-human interactions, perhaps the only setting where lies are considered acceptable or useful, if not outright necessary, in certain circumstances is the doctor-patient relationship. Indeed, this has been a topic of considerable intrigue in the medical community over the years. We thus end our paper with a brief discussion of the dynamics of white lies in the doctor-patient relationship in so much as it relates to the ethics of the design of human-AI interactions. 
We note that the following considerations also have strong cultural biases and some of these cultural artifacts are likely to feature in the characterization of an artificial agent’s behavior in different settings as well.

\subsubsection{The Hippocratic Oath}

Perhaps the strongest known support for deception in the practice of medicine is in the Hippocratic Decorum \cite{hippo} which states --

\vspace{5pt}
\noindent {\em
\textcolor{gray}{Perform your medical duties calmly and adroitly, \textcolor{black}{concealing most things from the patient while you are attending to him}. Give necessary orders with cheerfulness and sincerity, \textcolor{black}{turning his attention away from what is being done to him}; sometimes reprove sharply and sometimes comfort with solicitude and attention, \textcolor{black}{revealing nothing of the patient's future or present condition, for many patients through this course have taken a turn for the worse}.}
}

\vspace{5pt}
Philosophically, there has been no consensus \cite{bok1999lying} on this topic -- the Kantian view has perceived lies as immoral under all circumstances while the utilitarian view justifies the same ``greater good'' argument as put forward in our discussions so far. 
Specifically as it relates to clinical interactions, lies has been viewed variously from an impediment to treatment \cite{kernberg1985borderline} to a form of clinical aid. As Oliver Wendell Holmes put it 
\cite{holmes1892medical} -- 

\vspace{5pt}
\noindent {\em
``Your patient has no more right to all the truth you know than he has to all the medicine in your saddlebag\ldots he should only get just so much as is good for him.''}

\vspace{5pt}
The position we took on deception in the human-robot setting is similarly patronizing. It is likely to be the case that in terms of superior computational power or sensing capabilities there might be situations where the machine is capable of making decisions for the team that preclude human intervention but not participation. Should the machine be obliged to or even find use in revealing the entire truth in those situations? Or should we concede to our roles in such a relationship as we do with our doctors? 
This is also predicated on how competent the AI system is and to what extent it can be sure of the consequences \cite{hume1907essays} of its lies. This remains the primary concern for detractors of the ``greater goods'' doctrine, and the major deterrent towards the same.

\subsubsection{Root Causes of Deception in Clinical Interactions}

It is useful to look at the two primary sources of deception in clinical interactions 
-- (1) to hide mistakes (2) delivery of bad news \cite{palmieri2009lies}. The former is relevant to both the patient, who probably does not want to admit to failing to follow the regiment, and the doctor, who may be concerned about legal consequences. Such instances of deception to conceal individual fallibilities are out of scope of the current discussion. The latter scenario, on the other hand, comes from a position of superiority of knowledge about the present as well as possible outcomes in future, and has parallels to our current discussion. The rationale, here, being that such information can demoralize the patient and impede their recovery. It is interesting to note that the support for such techniques (both from the doctor’s as well as the patient’s perspectives) has decreased significantly over time \cite{wash}. That is not to say that human-machine interactions will be perceived similarly. As we saw in our study, participants were more or less open to the idea of deception or manipulation for greater good, especially in the event of a robotic teammate.

\subsubsection{Deception and Consent}

A related topic is, of course, that of consent -- if the doctor is not willing to reveal the whole truth, then what is the patient consenting to? In the landmark Slater vs Blaker vs Stapleton case (1767) \cite{annas2012doctors} the surgeon's intentions were indeed considered malpractice (the surgeon has broken the patient’s previously broken leg, fresh from a botched surgery, without consent and then botched the surgery again!). More recently, in the now famous Chester vs Afshar case (2004) \cite{cass2006nhs} the surgeon was found guilty of failing to notify even a 1-2\% chance of paralysis even though the defendant did not have to prove that they would have chosen not to have the surgery if they were given that information. In the context of human-machine interactions, it is hard to say then what the user agreement will look like, and whether there will be such a thing as consenting to being deceived, if only for the greater good, and what the legal outcomes of this will be when the interactions do not go as planned.

\subsubsection{The Placebo Effect}

Indeed, the effectiveness of placebo medicine, i.e. medicine prescribed while known to have no clinical effect, in improving patient symptoms is a strong argument in favor of deception in the practice of medicine. However, ethics of placebo treatment suggest that their use be limited to rare exceptions where \cite{hume1907essays} (1) the condition is known to have a high placebo response rate; (2) the alternatives are ineffective and/or risky; and (3) the patient has a strong need for some prescription. Further, the effectiveness of placebo is contingent on the patient’s trust on the doctor which is likely to erode as deceptive practices become common knowledge (and consequently render the placebo useless in the first place). Bok \cite{bok1999lying} points to this notion of ``cumulative harm''.

\subsubsection{Primum Non Nocere}

Perhaps the most remarkable nature of the doctor-patient relationship is captured by the notion of the recovery plot \cite{hak2000collusion} as part of a show being orchestrated by the doctor, and the patient being only complicit, while being cognizant of their specific roles in it, with the expectation of restoration of autonomy \cite{thomasma1994telling}, i.e. the state of human equality, free from the original symptoms or dependence on the doctor, at the end of the interaction. This is to say that the doctor-patient relationship is understood to be asymmetric and ``enters into a calculus of values wherein the respect for the right to truth of the patient is weighed against impairing the restoration of autonomy by the truth'' \cite{swaminath2008doctor} where the autonomy of the patient has historically taken precedence over beneficence and nonmalfeasance \cite{swaminath2008doctor}. 

\vspace{5pt}
\noindent In general, a human-machine relationship lacks this dynamic. So, while there are interesting lessons to be learned from clinical interactions with regards to value of truth and utility of outcomes, one should be carefully aware of the nuances of a particular type of relationship and situate an interaction in that context. Such considerations are also likely to shift according to the stakes on a decision, for example, lives lost in search and rescue scenarios. The doctor-patient relationship, and the intriguing roles of deception in it, does provide an invaluable starting point for conversation on the topic of greater good in human-AI interactions.

\subsection{Conclusions}

In this paper, we investigated the idea of fabrication, falsification and obfuscation of information when working with humans in the loop, and how such methods can be used by an AI agent to achieve teaming performance that would otherwise not be possible. This is increasingly likely to become an issue in the design of autonomous agents as AI agents become stronger and stronger in terms of computational and information processing capabilities thus faring better that their human counterparts in terms of cognitive load and situational awareness. We discussed how such behavior can be manufactured using existing AI algorithms, and used responses from participants in a thought experiment to gauge public perception on this topic.

The question of white lies and obfuscation or manipulation of information for the greater good is, of course, not unheard of in human-human interactions. A canonical example, as we saw in the final discussion, is the doctor-patient relationship where a doctor might have to withhold certain information to ensure that the patient has the best chance to recover, or might explain to the patient in different, and maybe simpler terms, than she would to a peer. It is unclear then how such behavior will be interpreted when attributed to a machine. We saw in the final case study that expectations and dynamics of a doctor-patient relation are very well-defined and do not necessarily carry over to a teaming setting. However, existing norms in doctor-patient relations do provide useful guidance towards answering some of the ethical questions raised by algorithms for greater good.

From the results of the survey presented in the paper, it seems that the public is, at least at the abstract level of the thought experiment, {\em positive towards lying for the greater good} especially when those actions would not be determined by their teammate, but is loath to suspend normative behavior, robot or not, in the event that they would be caught in that act {\em unless the robot is the recipient of the misinformation!} Further, most of the responses seem to be following a bimodal distribution indicating that the participants either felt strongly for or against this kind of behavior. 
It will be interesting to see if raising the stakes (for example, lives saved) of outcomes of these scenarios can contribute to a shift in perceived ethical consequences of such behavior, as seen in doctor-patient relationships. Another area that has seen evidences of AI being been used effectively to nudge human behavior is behavioral economics \cite{camerer2017artificial} which also raises similar interesting ethical dilemmas, and can be an interesting domain for further investigation. 

Finally, I note that all the use cases covered in the paper are, in fact, borne directly out of technologies or algorithms that I have developed \cite{chakraborti2015planning,ijcai2017-23}, albeit with slight modifications. as a student researcher over the last couple of years. Even though these algorithms were conceived with the best of intentions, such as to enable AI systems to explain their decisions or to increase effectiveness of collaborations with the humans in the loop, I would be remiss not to consider their ethical implications when used differently. In these exciting and uncertain times for the field of AI, it is thus imperative that researchers are cognizant of their scientific responsibility. I would like to conclude then by reiterating the importance of self-reflection in the principled design of AI algorithms whose deployment can have real-life consequences, intended or otherwise, on the future of the field, but also, with the inquisitive mind of a young researcher, marvel at the widening scope of interactions with an artificial agent into newer uncharted territories that may be otherwise considered to be unethical.

\bibliographystyle{aaai}
\bibliography{bib}

\end{document}